\documentclass{article}
\usepackage{arxiv}

\usepackage[thai, main=british]{babel}
\usepackage[utf8x]{inputenc} 
\usepackage[titletoc]{appendix}
\usepackage[T1]{fontenc}    
\usepackage{hyperref}       
\usepackage{url}            
\usepackage{booktabs}       
\usepackage{nicefrac}       
\usepackage{microtype}      
\usepackage{lipsum}

\usepackage{amsmath,amsfonts,amsthm}
\usepackage{cleveref}
\usepackage{pgfplots}
\usepackage{multirow}
\usepackage{caption}
\usepackage{subcaption}
\usepackage{tikz}
\usepackage{subfiles}
\usepackage{bbm}
\usepackage{authblk}

\renewcommand\appendix{\par
\setcounter{section}{0}%
\setcounter{subsection}{0}%
\gdef\thesection{\Alph{section}}}

\begin{document}

\title{Semi-supervised Thai Sentence Segmentation Using Local and Distant Word Representations}

\author[1]{Chanatip Saetia}
\author[1]{Ekapol Chuangsuwanich}
\author[2]{Tawunrat Chalothorn}
\author[1]{Peerapon Vateekul}

\affil[1]{Department of Computer Engineering\\
Chulalongkorn University\\
Bangkok, Thailand\\
\authorcr 6170135321@student.chula.ac.th, \{\tt ekapol.c, peerapon.v\}chula.ac.th
}
\affil[2]{
  Kasikorn Business Technology Group (KBTG)\\
  Nontaburi, Thailand \\
  \authorcr tawunrat.c@kbtg.tech \\
}



\maketitle

\begin{abstract}
  A sentence is typically treated as the minimal syntactic unit used for extracting valuable information from a longer piece of text. However, in written Thai, there are no explicit sentence markers. We proposed a deep learning model for the task of sentence segmentation that includes three main contributions. First, we integrate n-gram embedding as a local representation to capture word groups near sentence boundaries. Second, to focus on the keywords of dependent clauses, we combine the model with a distant representation obtained from self-attention modules. Finally, due to the scarcity of labeled data, for which annotation is difficult and time-consuming, we also investigate and adapt Cross-View Training (CVT) as a semi-supervised learning technique, allowing us to utilize unlabeled data to improve the model representations. In the Thai sentence segmentation experiments, our model reduced the relative error by 7.4\% and 10.5\% compared with the baseline models on the Orchid and UGWC datasets, respectively. We also applied our model to the task of pronunciation recovery on the IWSLT English dataset. Our model outperformed the prior sequence tagging models, achieving a relative error reduction of 2.5\%. Ablation studies revealed that utilizing n-gram presentations was the main contributing factor for Thai, while the semi-supervised training helped the most for English.
\end{abstract}

\keywords{Thai, Sentence segmentation, Punctuation restoration, Semi-supervised learning, Cross-View Training}

\section{Introduction}
Automatic summarization, machine translation, question answering, and semantic parsing operations are useful for processing, analyzing, and extracting meaningful information from text. However, when applied to long texts, these tasks usually require some minimal syntactic structure to be identified, such as sentences \cite{mihalcea2004graph, afsharizadeh2018query, raiman2017globally}, which always end with a period (``.'') in English \cite{gillick2009sentence}.

However, written Thai does not use an explicit end-of-sentence marker to identify sentence boundaries \cite{aroonmanakun2007thoughts}. Prior works have adapted traditional machine learning models to predict the beginning position of a sentence. The authors of \cite{mittrapiyanuruk2000automatic, charoenpornsawat2001automatic, slayden2010thai} proposed traditional models to determine whether a considered space is a sentence boundary based on the words and their part of speech (POS) near the space. Meanwhile, Zhou N. et al. \cite{zhou-etal-2016-word} considered Thai sentence segmentation as a sequence tagging problem and proposed a CRF-based model with n-gram embedding to predict which word is the sentence boundary. This method achieves the state-of-the-art result for Thai sentence segmentation and achieves greater accuracy than other models by approximately 10\% on an Orchid dataset \cite{sornlertlamvanich1997orchid}.

Several deep learning approaches have been applied in various tasks of natural language processing (NLP), including the long short-term memory \cite{hochreiter1997long}, self-attention \cite{vaswani2017attention}, and other models. Huang Z. et al. \cite{huang2015bidirectional} proposed a deep learning sequence tagging model called Bi-LSTM-CRF, which integrates a conditional random field (CRF) module to gain the benefit of both deep learning and traditional machine learning approaches. In their experiments, the Bi-LSTM-CRF model achieved an improved level of accuracy in many NLP sequence tagging tasks, such as named entity recognition, POS tagging and chunking.

The CRF module achieved the best result on the Thai sentence segmentation task \cite{zhou-etal-2016-word}; therefore, we adopt the Bi-LSTM-CRF model as our baseline. This paper makes the following three contributions to improve Bi-LSTM-CRF for sentence segmentation.

First, we propose adding n-gram embedding to Bi-LSTM-CRF due to its success in \cite{zhou-etal-2016-word} and \cite{huang2015bidirectional}. By including n-gram embedding, the model can capitalize on both approaches. First, the model gains the ability to extract past and future input features and sentence level tag information from Bi-LSTM-CRF; moreover, with the n-gram addition, it can also extract a local representation from n-gram embedding, which helps in capturing word groups that exist near sentence boundary. Although Jacovi A. et al. \cite{jacovi2018understanding} reported that a convolutional neural network (CNN) can be used as an n-gram detector to capture local features, we chose n-gram embedding over a CNN due to its better accuracy, as will be shown in Section~\ref{appendix:cnn}.

Second, we propose adding incorporative distant representation into the model via a self-attention mechanism, which can focus on the keywords of dependent clauses that are far from the considered word. Self-attention has been used in many recent state-of-the-art models, most notably the transformer \cite{vaswani2017attention} and BERT \cite{devlin2018bert}. BERT has outperformed Bi-LSTM on numerous tasks, including question answering and language inference. Therefore, we choose to use self-attention modules to extract distant representations along with local representations to improve model accuracy.

Third, we also apply semi-supervised learning \cite{zhu2005semi}, allowing us to employ unlimited amounts of unlabeled data, which is particularly important for low-resource languages such as Thai, for which annotation is costly and time-consuming. Many semi-supervised learning approaches have been proposed in the computer vision \cite{rasmus2015semi, miyato2018virtual} and natural language processing \cite{miyato2016adversarial, ruder2018strong, clark2018semi} fields. Our choice for semi-supervised learning to enhance model representation is Cross-View Training (CVT) \cite{clark2018semi}. Clark K. et al. \cite{clark2018semi} claims that CVT can improve the representation layers of the model, which is our goal. However, CVT was not designed to be integrated with self-attention and CRF modules; consequently, we provide a modified version of CVT in this work.

Based on the above three contributions, we pursue two main experiments. The first experiment was conducted on two Thai datasets, Orchid and UGWC \cite{lertpiya2019preliminary}, to evaluate our Thai sentence segmentation model. In this case, our model achieves F1 scores of 92.5\% and 88.9\% on Orchid and UGWC, respectively, and it outperforms all the baseline models. The second experiment was executed on the IWSLT dataset \cite{federico2012overview} and involves an English-language punctuation restoration task. This experiment demonstrates that our model is generalizable to different languages. Our model, which does not require pretrained word vectors, improved the overall F1 score by 0.9\% compared to the baselines, including a model that uses pretrained word vectors.

There are five sections in the remainder of this paper. Section~\ref{sec:related} reviews the related works on Thai sentence segmentation, English punctuation restoration and introduces the original CVT. Section~\ref{sec:method} describes the proposed model architecture and the integration of cross-view training. The datasets, implementation process and evaluation metrics are explained in Section~\ref{sec:experiment}. The results of the experiments are discussed in Section~\ref{sec:result}. Finally, Section~\ref{sec:conclusion} concludes the paper.

\section{Related Works} \label{sec:related}

This section includes three subsections. The first subsection concerns Thai sentence segmentation, which is the main focus of this work. The task of English punctuation restoration, which is similar to our main task, is described in the second subsection. The last subsection describes the original Cross-View Training initially proposed in \cite{clark2018semi}.

\subsection{Thai sentence segmentation}
In Thai, texts do not contain markers that definitively identify sentence boundaries. Instead, written Thai usually uses a space as the vital element that separates text into sentences. However, there are three ways that spaces are used in this context \cite{wathabunditkul2003spacing}: before and after an interjection, before conjunctions, and before and after numeric expressions. Therefore, segmenting text into sentences cannot be performed simply by splitting a text at the spaces.

Previous works from \cite{mittrapiyanuruk2000automatic, charoenpornsawat2001automatic, slayden2010thai} have focused on disambiguating whether a space functions as the sentence boundary. These works extract contextual features from words and POS around the space. Then, the obtained features around the corresponding space are input into traditional models to predict whether space is a sentence boundary.

Although a space is usually considered essential as a sentence boundary marker, approximately 23\% of the sentences end without a space character in one news domain corpus \cite{zhou-etal-2016-word}. Hence, Zhou N. et al. \cite{zhou-etal-2016-word} proposed a word sequence tagging CRF-based model in which all words can be considered as candidates for the sentence boundary. A space is considered as only one possible means of forming a sentence boundary. The CRF-based model \cite{lafferty2001conditional}, which is extracted from n-grams around the considered word, achieves a F1 score of 91.9\%, which is approximately 10\% higher than the F1 scores achieved by other models \cite{mittrapiyanuruk2000automatic, charoenpornsawat2001automatic, slayden2010thai} on the Orchid dataset, as mentioned in \cite{zhou-etal-2016-word}.

In this work, we adopt the concept of word sequence tagging and compare it with two baselines: the CRF-based model with n-gram embedding, which is currently the state-of-the-art for Thai sentence segmentation, and the Bi-LSTM-CRF model, which is currently the deep learning state-of-the-art approach for sequence tagging.

\subsection{English punctuation restoration}
Most languages use a symbol that functions as a sentence boundary; however, a few do not use sentence markers including Thai, Lao and Myanmar. Thus, few studies have investigated sentence segmentation in raw text. However, studies on sentence segmentation, which is sometimes called sentence boundary detection, are still found in the speech recognition field \cite{gotoh2000sentence}. The typical input to speech recognition model is simply a stream of words. If two sentences are spoken back to back, by default, a recognition engine will treat it as one sentence. Thus, sentence boundary detection is also considered a punctuation restoration task in speech recognition because when the model attempts to restores the period in the text, the sentence boundary position will also be defined.

Punctuation restoration not only provides a minimal syntactic structure for natural language processing, similar to sentence boundary detection but also dramatically improves the readability of transcripts. Therefore, punctuation restoration has been extensively studied. Many approaches have been proposed for punctuation restoration that use different features, such as audio and textual features. Moreover, punctuation restoration is also considered to be a different machine learning problem, such as word sequence tagging and machine translation.

A combination of audio and textual features were utilized in \cite{tilk2015lstm, kolavr2006using, kolavr2012development} to predict and restore punctuation, including pitch, intensity and pause duration, between words. We ignore these features in our experiment because our main task—Thai sentence segmentation— does not include audio features.

Focusing only on textual features, there are two main approaches, namely, word sequence tagging and machine translation. For the machine translation approach, punctuation is treated as just another type of token that needs to be recovered and included in the output. The methods in \cite{peitz2011modeling,cho2015punctuation,wang2018self} restore punctuation by translating from unpunctuated text to punctuated text. However, our main task, sentence segmentation, is an upstream task in text processing, unlike punctuation restoration, which is considered a downstream task. Therefore, the task needs to operate rapidly; consequently, we focus only on the sequence tagging model, which is less complex than the machine translation model.

In addition to those machine translation tasks, both traditional approaches and deep learning approaches must solve a word sequence tagging problem. Of the traditional approaches, contextual features around the considered word were used to predict following punctuation in the n-gram \cite{gravano2009restoring} and CRF model approaches \cite{lu2010better, ueffing2013improved}. Meanwhile, in the deep learning approaches, a deep convolutional neural network \cite{che2016punctuation}, T-LSTM (Textual-LSTM) \cite{tilk2015lstm} and a bidirectional LSTM model with an attention mechanism, called T-BRNN \cite{tilk2016bidirectional}, have been adopted to predict a punctuation sequence from the word sequence. T-BRNN \cite{tilk2016bidirectional} was proposed to solve the task as a word-sequence tagging problem, and it is currently the best model that uses the word sequence tagging approach. Tilk O. et al. \cite{tilk2016bidirectional} also proposed a variant named T-BRNN-pre, which integrates pretrained word vectors to improve the accuracy.

To demonstrate that our model is generalizable to other languages, we compare it with other punctuation restoration models, including T-LSTM, T-BRNN, and T-BRNN-pre. These models adopt a word sequence tagging approach and do not utilize any prosodic or audio features.

\subsection{Cross-View Training} \label{sec:related_cvt}

CVT \cite{clark2018semi} is a semi-supervised learning technique whose goal is to improve the model representation using a combination of labeled and unlabeled data. During training, the model is trained alternately with one mini-batch of labeled data and $B$ mini-batches of unlabeled data.

Labeled data are input into the model to calculate the standard supervised loss for each mini-batch and the model weights are updated regularly. Meanwhile, each mini-batch of unlabeled data is selected randomly from the pool of all unlabeled data; the model computes the loss for CVT from the mini-batch of unlabeled data. This CVT loss is used to train auxiliary prediction modules, which see restricted views of the input, to match the output of the primary prediction module, which is the full model that sees all the input. Meanwhile, the auxiliary prediction modules share the same intermediate representation with the primary prediction module. Hence, the intermediate representation of the model is improved through this process.

Similar to the previous work, we also apply CVT to a sequence tagging task. However, our model is composed of self-attention and CRF modules, which were not included in the sequence tagging model in \cite{clark2018semi}. The previous CVT was conducted on an LSTM using the concepts of forward and backward paths, which are not intuitively acquired by the self-attention model. Moreover, the output used to calculate CVT loss was generated by the softmax function, which does not operate with CRF. Thus, in our study, both the primary and auxiliary prediction modules needed to be constructed differently from the original ones.

\section{Proposed method} \label{sec:method}
In this section, we describe our proposed method in two subsections. The first subsection specifies the model architecture and the details of each module. Our first and second contributions, which are local and distant representations, are mainly described in this subsection. Meanwhile, the second subsection expounds on how the model is trained with unlabeled data through the modified CVT, which is our third contribution.

\subsection{Model architecture}

In this work, the model predicts the tags $\vec{y}=\left[y_1,y_2,\ \ldots,\ y_N\right],\ \forall y\in Y$ for the tokens in a word sequence $\vec{x}=\left[x_1,x_2,\ \ldots,\ x_N\right]$ where $N$ is the sequence size and $x_t$, $y_t$ denote the token and its tag at timestep $t$, respectively. Each token $x_t$ consists of a word, its POS and its type. There are five defined word types: English, Thai, punctuation, digits, and spaces.

The tag set $Y$ is populated based on the considered task. In Thai sentence segmentation, the assigned tags are $sb$ and $nsb$; $sb$ denotes that the corresponding word is a sentence boundary considered as the beginning of a sentence, while and $nsb$ denotes that the word is not a sentence boundary. Meanwhile, there are four tags in the punctuation restoration task. Words not followed by any punctuation are tagged with $O$. Words that are followed by a period ``.'', comma ``,'' or question mark ``?'' are tagged to $PERIOD$, $COMMA$, and $QUESTION$, respectively.

\begin{figure}
	\centering
\begin{tikzpicture}
\tikzstyle{lowmodule} = [rectangle, minimum width=7.5cm, minimum height=3cm, draw=black]
\tikzstyle{highmodule} = [rectangle, minimum width=7.5cm, minimum height=2.2cm, draw=black]
\tikzstyle{predictionmodule} = [rectangle, minimum width=7.5cm, minimum height=2.6cm, draw=black]
\tikzstyle{smallnetwork} = [rectangle, minimum width=3.3cm, minimum height=1.2cm,
draw=black]
\tikzstyle{network} = [rectangle, minimum width=3.3cm, minimum height=2cm,
draw=black]
\tikzstyle{module} = [rectangle, minimum width=7cm, minimum height=0.6cm,
draw=black]
\tikzstyle{submodule} = [rectangle, minimum width=3cm, minimum height=0.6cm,
draw=black]
\tikzstyle{bisubmodule} = [rectangle, minimum width=3.3cm, minimum height=0.6cm,
draw=black]
\tikzstyle{2submodule} = [rectangle, minimum width=3cm, minimum height=1.35cm,
draw=black]
\tikzstyle{edge from parent}=[]

\foreach \text \x in {I/5.5, am/6.5, studying/7.5, NLP/8.5, are/9.5, n't/10.5, you/11.5}
{\node at (\x, 0.25) [anchor=north, text height=0.25] {\text};}

\node at (0.25, 0.75) [anchor=north west, text height=0.25] {Word token ($\vec{x}$)};

\foreach \x in {5.5,...,11.5}
{\draw [-latex] (\x,0.5) -- (\x,1);}

\node at (8.5, 2.5) (low) [lowmodule] {}
	child {node at (1,0) [anchor=south west] {Low-level module}};
\node at (6.75, 2.3) (local) [smallnetwork] {}
	child {node at (-0.7,0.9) [anchor=south west] {Local structure}};
\node at (6.75, 2.5) (ngram) [submodule] {Ngram embedding};
\node at (6.75, 3.4) (bilstm) [bisubmodule] {Bi-LSTM};
\node at (10.25, 2.7) (distant) [network] {}
	child {node at (-0.95,0.5) [anchor=south west] {Distant structure}};
\node at (10.25, 2.85) (lowatten) [2submodule] {Self-attention};

\node at (0.25, 4.25) [anchor=north west, text height=0.25] {Low-level representation ($\vec{r}$)};

\foreach \x in {5.5,...,11.5}
{\draw [-latex] (\x,4) -- (\x,4.5);}

\node at (8.5, 5.6) (high) [highmodule] {}
	child {node at (0.9,0.35) [anchor=south west] {High-level module}};
\node at (8.50, 5.35) (stack) [module] {Stack Bi-LSTM};
\node at (8.50, 6.15) (highatten) [module] {Self-attention};

\node at (0.25, 6.95) [anchor=north west, text height=0.25] {High-level representation ($\vec{h}$)};

\foreach \x in {5.5,...,11.5}
{\draw [-latex] (\x,6.70) -- (\x,7.2);}

\node at (8.5, 8.5) (predict) [predictionmodule] {}
	child {node at (0.9,0.25) [anchor=south west] {Prediction module}};
\node at (8.50, 8.1) (stack) [module] {Fully Connected};

\node at (0.25, 8.65) [anchor=north west, text height=0.25] {Virtual logit ($\vec{g}$)};

\foreach \x in {5.5,...,11.5}
{\draw [-latex] (\x, 8.4) -- (\x,8.9);}

\node at (8.50, 9.2) (highatten) [module] {CRF};

\node at (0.25, 10.05) [anchor=north west, text height=0.25] {Prediction ($\vec{y}$)};

\foreach \x in {5.5,...,11.5}
{\draw [-latex] (\x,9.8) -- (\x,10.3);}

\foreach \text \x in {O/5.5, O/6.5, O/7.5, comma/8.5, O/9.5, O/10.5, question/11.5}
{\node at (\x, 10.55) [anchor=north, text height=0.25] {\text};}

\end{tikzpicture}
  \caption{Model architecture that integrates local and distant representation. This model is composed of three main modules: a low-level module, a high-level module and a prediction module. In the low-level module, two structures (local and distant) are responsible for extracting different features.}
  \label{fig:architecture}
\end{figure}
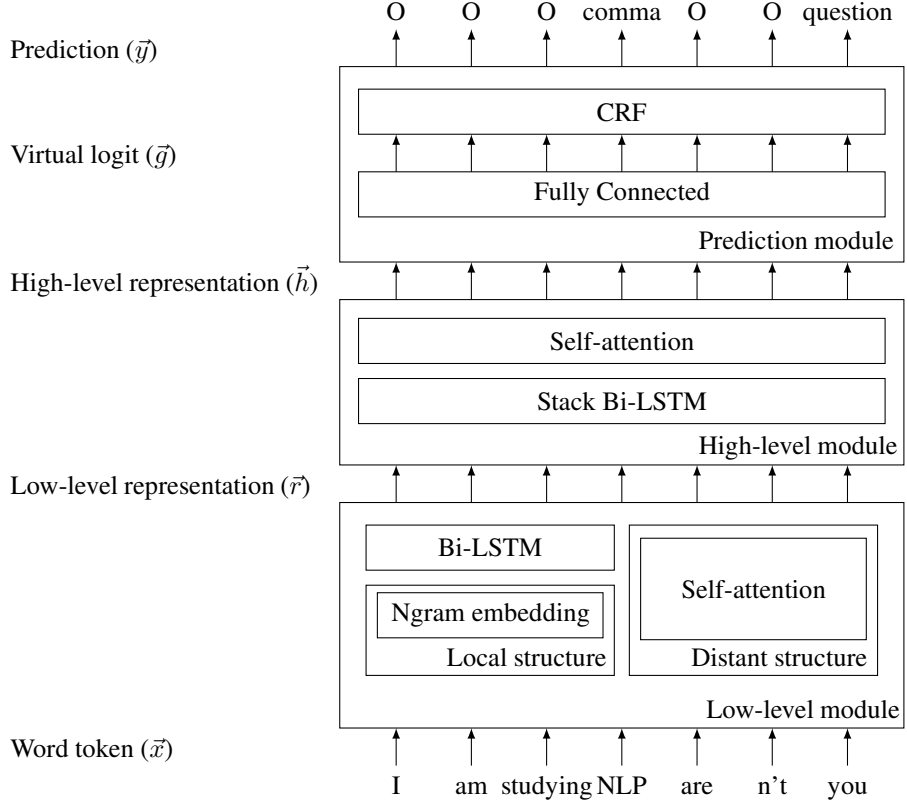

Our model architecture is based on Bi-LSTM-CRF, as shown in Fig.~\ref{fig:architecture}. The model is divided into three modules. The first, low-level module, consists of two separate structures: local and distant structures. The second, high-level module, contains a sequence of stacked bidirectional LSTM and self-attention layers. The final module, the prediction module, is responsible for predicting the tags $\vec{y}$. Each module is described more completely in the next three subsections.

\subsubsection{Low-level module} \label{sec:low-level}
A sequence of word tokens is input into the low-level module. The input tokens pass through two structures. The first structure generates a sequence of local representation vectors $\mathbf{R}_{local}=[\vec{r}_{1,local},\vec{r}_{2,local},\ldots,\vec{r}_{N,local}]$, and the second structure generates low-level distant representation vectors $\mathbf{R}_{distant}=[\vec{r}_{1,distant},\vec{r}_{2,distant},\ldots,\vec{r}_{N,distant}]$. After obtaining both sequences of representation vectors, the local representation vectors are fed to the Bi-LSTM to obtain the recurrent representation vectors $\mathbf{R}_{recurrent}=[\vec{r}_{1,recurrent},\vec{r}_{2,recurrent},\ldots,\vec{r}_{N,recurrent}]$, as shown in \Cref{recurrent_rep}. Then, the recurrent and distant representation vectors are concatenated to form the low-level representation vector $\mathbf{R}=[\vec{r}_1,\vec{r}_2,\ldots,\vec{r}_N]$, as shown in \Cref{low-level}:

\begin{equation}
  \mathbf{R}_{recurrent} = \operatorname{BiLSTM} ( \mathbf{R}_{local} )
  \label{recurrent_rep}
\end{equation}
\begin{equation}
  \vec{r}_t =  \vec{r}_{t, recurrent} \oplus \vec{r}_{t,distant}
  \label{low-level}
\end{equation}

\paragraph{Local structure}
This structure is shown as the left submodule of the low-level module in Fig.~\ref{fig:architecture}. It extracts the local representation vectors $\mathbf{R}_{local}$. Its input tokens are used to create n-gram tokens, which are unigrams $x_a$, bigrams $(x_a, x_b)$, and trigrams $(x_a, x_b, x_c)$. Each n-gram token is represented as an embedding vector, which is classified as a unigram embedding vector ${\vec{e}}_{uni}$, a bigram embedding vector ${\vec{e}}_{bi}$ or a trigram embedding vector ${\vec{e}}_{tri}$. Each vector ${\vec{e}}_{gram}$ is mapped from a token by gram embedding $\operatorname{Embedding}_{gram}\left(x\right)$, which is a concatenated vector of the word embedding $\operatorname{Word_{gram}}\left(x\right)$, POS embedding $\operatorname{POS_{gram}}\left(x\right)$ and type embedding $\operatorname{Type_{gram}}\left(x\right)$, as shown in \Cref{eq:embedding}:

\begin{equation}
  \operatorname{Embedding_{gram}}( x ) = \operatorname{W_{gram}}( x ) \oplus \operatorname{POS_{gram}}( x ) \oplus \operatorname{Type_{gram}}( x )
  \label{eq:embedding}
\end{equation}

Each n-gram token at timestep $t$ is generated by the previous, present and next token and embedded into vectors as shown in ~\Cref{uni,bi,tri}. The unigram embedding at timestep $t$ is a unigram embedding of the current token $x_t$. The bigram embedding vector at timestep $t$ is a bigram embedding of the previous and present tokens $(x_{t-1}, x_t)$, and the trigram embedding vector at timestep $t$ is a trigram embedding of the previous, present and next tokens $(x_{t-1}, x_t,x_{t+1})$:

\begin{equation}
 \vec{e}_{t,uni} = \operatorname{Embedding_{uni}}( x_{t} )
  \label{uni}
\end{equation}
\begin{equation}
 \vec{e}_{t,bi} = \operatorname{Embedding_{bi}}( x_{t-1} , x_{t} )
  \label{bi}
\end{equation}
\begin{equation}
 \vec{e}_{t,tri} = \operatorname{Embedding_{tri}}( x_{t-1} , x_{t} , x_{t+1} )
  \label{tri}
\end{equation}

At each timestep $t$, a local representation vector $\vec{r}_{t,local}$ is combined from the n-gram embedding vectors generated from the context around $x_t$. A combination of embedding vectors, which is used to construct a local representation vector, is shown in \Cref{n-gram_combination}. A combination consists of the unigram, bigram, and trigram embedding vectors at timesteps $t-1$, $t$ and $t+1$ and it is a concatenation of all the embedding vectors:

\begin{equation}
  \vec{r}_{t,local} = \vec{e}_{t-1, uni} \oplus \vec{e}_{t,uni} \oplus \vec{e}_{t+1,uni} \oplus 
  \vec{e}_{t-1,bi} \oplus \vec{e}_{t,bi}\oplus \vec{e}_{t+1,bi}\oplus 
  \vec{e}_{t-1,tri} \oplus  \vec{e}_{t,tri} \oplus  \vec{e}_{t+1,tri}
  \label{n-gram_combination} 
\end{equation}

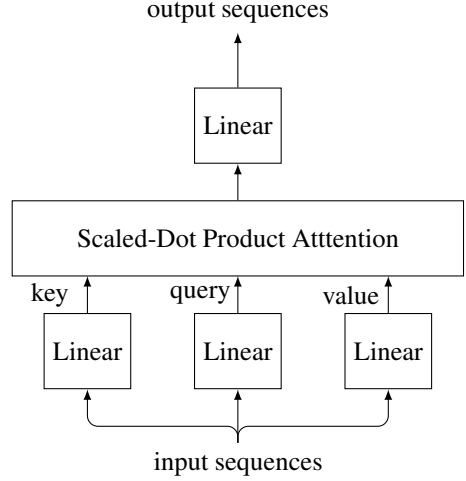
\begin{figure}
    \centering
    \begin{tikzpicture}
        \node[text centered] at (7, 0) (inpseq) {input sequences};
        \node[text centered,rectangle, minimum width=1cm, minimum height=1cm,
        draw=black] at (5, 1.5) (K) {Linear};
        \node[text centered,rectangle, minimum width=1cm, minimum height=1cm,
        draw=black] at (7, 1.5) (Q) {Linear};
        \node[text centered,rectangle, minimum width=1cm, minimum height=1cm,
        draw=black] at (9, 1.5) (V) {Linear};
        {\draw[-latex,rounded corners] (inpseq) -- ++(0,0.5) -| (K);}
        {\draw[-latex,rounded corners] (inpseq) -- (Q);}
        {\draw[-latex,rounded corners] (inpseq) -- ++(0,0.5) -| (V);}
        \node[text centered] at (4.5, 2.25) {key};
        \node[text centered] at (6.5, 2.25) {query};
        \node[text centered] at (8.5, 2.25) {value};
        \draw [-latex] (K) -- (5, 2.5);
        \draw [-latex] (Q) -- (7, 2.5);
        \draw [-latex] (V) -- (9, 2.5);
        \node[text centered,rectangle, minimum width=6cm, minimum height=1cm,
        draw=black] at (7, 3) (attention) {Scaled-Dot Product Atttention};
        \node[text centered,rectangle, minimum width=1cm, minimum height=1cm,
        draw=black] at (7, 4.5) (output) {Linear};
        \draw [-latex] (attention) -- (output);
        \node[text centered] at (7, 6) (outseq) {output sequences};
        \draw [-latex] (output) -- (outseq);
    \end{tikzpicture}
\caption{The architecture of a self-attention module. This module mainly contains Scaled-Dot Product Attention, which requires three inputs: Key, Query and Value. Those inputs are generated from the same input sequence but projected by different linear transformations.}

\label{fig:selfattention}
\end{figure}

\paragraph{Distant structure}
The distant structure, which is a self-attention module, is shown in Fig.~\ref{fig:architecture} on the right side of the low-level module. The structure extracts low-level distant representation vectors $\mathbf{R}_{distant}$ from a sequence of unigram embedding vectors $\mathbf{E}_{uni}$, as shown in \Cref{distant_attention}. In this case, the self-attention module is a scaled dot-product attention \cite{vaswani2017attention}, where key, query, and value vectors are the linear projections of the unigram embedding vectors shown in Fig.~\ref{fig:selfattention}. The linear transformations for key, query, and value are learned separately and updated in the model through backpropagation. The output vector, which is the scaled dot-product attention at each timestep, is concatenated with the input vector ${\vec{e}}_{t,uni}$ and projected by a linear transformation. That projected vector is the output vector of a self-attention module, which is a low-level distant representation vector.

\begin{equation}
  \mathbf{R}_{distant} = \operatorname{SelfAttention} ( \mathbf{E}_{uni} )
  \label{distant_attention}
\end{equation}

\subsubsection{High-level module} \label{sec:high-level}
The low-level representation vectors $\mathbf{R}$ are used as the input for this module, which outputs the high-level representation vectors $\mathbf{H}$ whose calculation is shown in \Cref{high-level}. The high-level module, as shown in Fig.~\ref{fig:architecture}, is composed of a stacked bidirectional LSTM and a self-attention modules. A stacked bidirectional LSTM contains K layers of bidirectional LSTMs in which the output from the previous bidirectional LSTM layer is the input of the next bidirectional LSTM layer. The self-attention part of this structure is the same as that in the low-level distant structure. The self-attention module helps to generate the high-level distant representation vectors that are output by the high-level module.

\begin{equation}
  \mathbf{H} = \operatorname{SelfAttention} (  \operatorname{StackBiLSTM} ( \mathbf{R} )  )
  \label{high-level}
\end{equation}

\subsubsection{Prediction module}
The prediction module is the last module. It includes two layers: a fully connected layer and a CRF layer. In the fully connected layer, the output vectors from the high-level module are projected by a linear transformation as shown in \Cref{virtual}. The purpose of this layer is to create the virtual logit vectors $\mathbf{G} = [\vec{g}_1, \vec{g}_2, ..., \vec{g}_N]$, which represent the probability distribution for CVT, as discussed in Section~\ref{cross-view}. Therefore, the number of dimensions of logits equals the number of possible tags in each task:

\begin{equation}
  \vec{g}_t = \operatorname{NN}( \vec{h}_t ) 
  \label{virtual}
\end{equation}

The CRF layer is responsible for predicting the tag $y_t$ of a token at each timestep, as shown in \Cref{crf}. The layer receives a sequence of virtual logit vectors ($\mathbf{G}$) as input and then decodes them to a sequence of tags $\vec{y}$ using the Viterbi algorithm.

\begin{equation}
  \vec{y} = \operatorname{CRF}( \mathbf{G} ) 
  \label{crf}
\end{equation}

\subsection{Cross-View Training}  \label{cross-view}


\tikzstyle{lowmodule} = [rectangle, minimum width=7.2cm, minimum height=1.9cm, draw=black]
\tikzstyle{network} = [rectangle, minimum width=3.3cm, minimum height=0.6cm,
draw=black]
\tikzstyle{distantnetwork} = [rectangle, minimum width=3.3cm, minimum height=1.4cm,
draw=black]
\tikzstyle{module} = [rectangle, minimum width=7.5cm, minimum height=0.6cm,
draw=black]
\tikzstyle{edge from parent}=[]

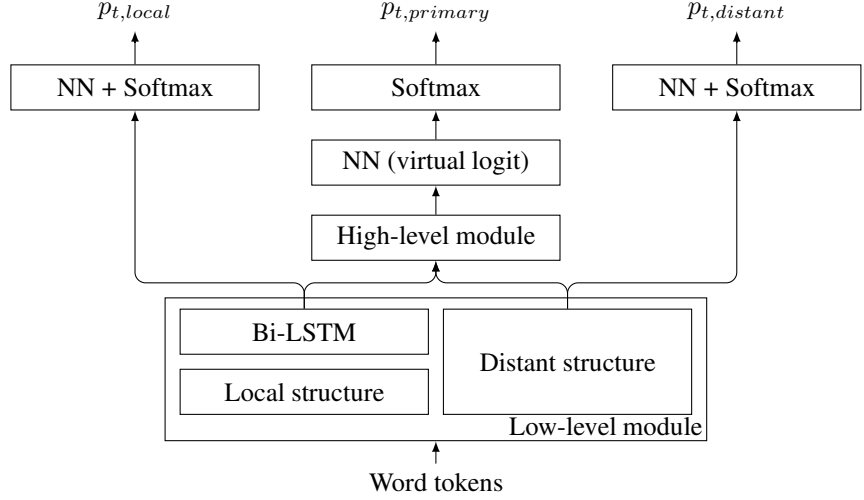
\begin{figure}
  \centering
    \begin{tikzpicture}
      \node at (7.5, 1.25) (low) [lowmodule] {}
          child {node at (0.85,0.5) [anchor=south west] {Low-level module}};
      

      \node at (7.5, 6) (primary) {$p_{t,primary}$};
      \node at (3.5, 6) (localaux) {$p_{t,local}$};
      \node at (11.5, 6) (globalaux) {$p_{t,distant}$};


      \node at (3.5, 5) (localnn) [network] {NN + Softmax};
      \node at (11.5, 5) (globalnn) [network] {NN + Softmax};
      \node at (7.5, 5) (primarynn) [network] {Softmax};
      \node at (5.75, 0.95) (local) [network] {Local structure};
      \node at (5.75, 1.75) (lstm) [network] {Bi-LSTM};
      \node at (9.25, 1.35) (global) [distantnetwork] {Distant structure};
      \node at (7.5, 4) (logit) [network] { NN (virtual logit) }  edge [-latex] (primarynn);
      \node at (7.5, 3) (high) [network] {High-level module} edge [-latex] (logit);

      \node at (7.5, -0.25) (input) {Word tokens} edge [-latex] (low);

      \draw[-latex,rounded corners] (lstm) -- ++(0,0.65) -| (localnn);
      \draw[-latex,rounded corners] (global) -- ++(0,1.05) -| (globalnn);

      \draw[-latex,rounded corners] (primarynn) -- ++(0,0.5) -| (primary);
      \draw[-latex,rounded corners] (localnn) -- ++(0,0.5) -| (localaux);
      \draw[-latex,rounded corners] (globalnn) -- ++(0,0.5) -| (globalaux);


      \draw[-latex,rounded corners] (lstm) -- ++(0,0.65) -| (high);
      \draw[-latex,rounded corners] (global) -- ++(0,1.05) -| (high);

    \end{tikzpicture}
\caption{
  Two auxiliary predictions (${\vec{p}}_{t,local}$ and ${\vec{p}}_{t,distant}$) are obtained from the local and global structures in the low-level module. The primary prediction ${\vec{p}}_{t,primary}$ is obtained from the virtual logit vector ${\vec{g}}_t$}

\label{fig:CVT}
\end{figure}

As discussed in Section~\ref{sec:related_cvt}, CVT requires primary and auxiliary prediction modules for training with unlabeled data to improve the representation. Thus, we construct both types of prediction modules for our model. The flow of unlabeled data, which is processed to obtain a prediction by each module, is shown in Fig.~\ref{fig:CVT}. The output of each prediction module is transformed into the probability distribution of each class by the softmax function and then used to calculate ${Loss}_{CVT}$, as shown in \Cref{cvt_loss}.
 
\begin{equation}
  Loss_{CVT} = \frac{1}{|D|} \sum_{t \in D} D_{\operatorname{KL}}(\vec{p}_{t,primary}, \vec{p}_{t,local}) + D_{\operatorname{KL}}(\vec{p}_{t,primary}, \vec{p}_{t,distant})
  \label{cvt_loss}
\end{equation}

The ${Loss}_{CVT}$ value is based on the Kullback–Leibler divergence (KL divergence) between the probability distribution of the primary ${\vec{p}}_{t,primary}$ output and those of two auxiliary modules, ${\vec{p}}_{t,local}$ and ${\vec{p}}_{t,distant}$, where $t\in[1,...,N]$. The KL divergence at each timestep is averaged when the timesteps are dropped timesteps D, which is described in Section~\ref{sec:aux}.
The details of the primary and auxiliary prediction modules, which are used in the ${Loss}_{CVT}$ calculation, are described in the following subsections.

\subsubsection{Primary prediction module}
In \cite{clark2018semi}, the output of the primary prediction module is acquired from the last layer and used to predict tags. However, our model uses a CRF layer to decode the tags, instead of the softmax function, whose input is the output from the last fully connected layer. Thus, the probability distribution of a primary prediction module should be the marginal probability acquired from the CRF layer. Nevertheless, the forward-backward algorithm for the marginal probability calculation is time-consuming with a time complexity is $O\left(S^2N\right)$, where $N$ is the sequence length, and $S$ is the number of tags. To reduce the training time, the probability distribution of the primary prediction module ${\vec{p}}_{t,primary}$ is instead obtained from the output of the Softmax function, whose input is a virtual logit vector ${\vec{g}}_t$, as shown in \Cref{primary}.

\begin{equation}
\vec{p}_{t,primary} = \operatorname{Softmax}( \vec{g}_t )
\label{primary}
\end{equation}

\subsubsection{Auxiliary prediction module}  \label{sec:aux}
Two auxiliary views are included to improve the model. The first view is generated from a recurrent representation vector $r_{t,recurrent}$ to acquire the local probability distribution ${\vec{p}}_{t,local}$, where $t\in[1,...,N]$. The second view is generated from the low-level distant representation vectors $r_{t,distant}$ to acquire the probability distribution of a distant structure in the low-level module ${\vec{p}}_{t,distant}$, where $t\in[1,...,N]$. By generating the views from these representation vectors separately, the local and distant structures in the low-level module can improve equally.

Although both representation vectors are used separately to create auxiliary views, the input of each structure is still not restricted, unlike \cite{clark2018semi}, where the input is restricted to only previous or future tokens. Because BERT, which is trained by the masked language model, outperforms OpenAI GPT, which uses an autoregressive approach for training as reported in \cite{devlin2018bert}, we adopt the concept of the masked language model \cite{taylor1953cloze} to obtain both auxiliary views. This approach allows the representation to fuse the left and the right context, which results in a better representation. By using the masked language model, some tokens at each timestep are randomly dropped and denoted as removed tokens $\langle REMOVED \rangle$; then, the remaining tokens are used to obtain auxiliary predictions in the dropped timesteps $D = \{ d \in N | d \text{ is a dropped timestep} \} $, as shown in Fig.~\ref{fig:masklanguage}. The details of both auxiliary prediction modules are described below.

\tikzstyle{network} = [rectangle, minimum width=10cm, minimum height=1cm,
draw=black]

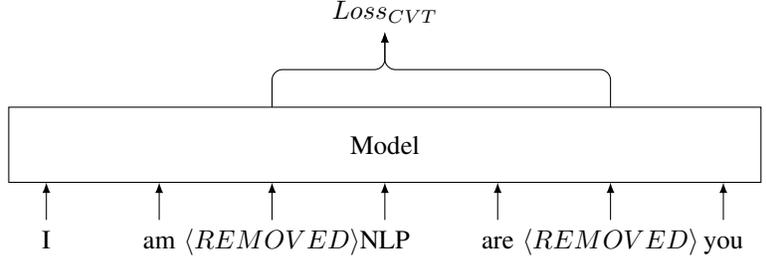
\begin{figure}
  \centering
    \begin{tikzpicture}
        \node at (7, 7) [network] (model) {Model};
        \foreach \text \x in {I/1, am/2, $\langle REMOVED \rangle$/3, NLP/4, are/5, $\langle REMOVED \rangle$/6, you/7}
          {\draw[-latex] (1 + \x * 1.5,6) -- (1 + \x * 1.5,6.5);
          \node at (1 + \x * 1.5, 5.75) [anchor=north, text height=0.25] {\text};}
          
        \foreach \x in {3,6}
          {\draw[-latex,rounded corners] (1 + \x * 1.5,7.5) -- ++(0,0.5) -| (7,8.5);}
        \node[text centered, anchor=south] at (7, 8.5) (output) {$Loss_{CVT}$};
    \end{tikzpicture}
\caption{
  Words are dropped (denoted as $\langle REMOVED \rangle$) randomly. Those positions are used to calculate $Loss_{CVT}$ and update the auxiliary prediction modules to improve the model. 
}
\label{fig:masklanguage}
\end{figure}

\paragraph{Local auxiliary module}
For recurrent representation vectors, if one of the tokens is dropped, the related n-gram tokens that include the dropped tokens will also be dropped. For example, if $\left(x_t\right)$ is dropped, $\left(x_{t-1},x_t\right)$ and $\left(x_t,x_{t+1}\right)$ will also be dropped as removed tokens in the case of a bigram. The remaining n-gram tokens are then used to obtain the recurrent representation vectors at the dropped timesteps. Then, the vectors are provided as an input to the softmax function to obtain the probability distribution of the first auxiliary prediction module, as shown in \Cref{aux_local}.

\begin{equation}
\vec{p}_{d,local} = \operatorname{Softmax}( \operatorname{NN} (\vec{r}_{d,recurrent} ) )
\label{aux_local}
\end{equation}

\paragraph{Distant auxiliary module}
In the other auxiliary prediction module, a sequence of the low-level distant representation vectors is generated and some tokens are dropped. This sequence of vectors is also input into the Softmax function, just as in the first auxiliary prediction module, and the output is another probability distribution, which is the second auxiliary prediction, as shown in \Cref{aux_distant}.

\begin{equation}
\vec{p}_{d,distant} = \operatorname{Softmax}( \operatorname{NN} (\vec{r}_{d,distant} ) )
\label{aux_distant}
\end{equation}

\section{Experimental setup} \label{sec:experiment}
\subsection{Datasets}

Three datasets are used in the experiments as described in the following subsections.
We use two datasets for Thai sentence segmentation, and the third dataset is used for English punctuation restoration. The statistics of the preprocessed data are shown in Table~\ref{tab:vocab}, including the number of sequences and the number of vocabulary words in each dataset. We also calculate the average number of words per passage in the unlabeled data that do not appear in the labeled data, as shown in Table~\ref{tab:new_vocab}.

\begin{table}
  \caption{The number of passages and vocabulary words in each dataset. The labeled and unlabeled data are separately counted and shown in the rows.}
  \centering
  \label{tab:vocab}
  \begin{tabular}{p{4cm}|cc|cc|cc}
    \toprule
    Dataset & \multicolumn{2}{c|}{Orchid (Thai)} & \multicolumn{2}{c|}{UGWC (Thai)} & \multicolumn{2}{c}{IWSLT (Eng)} \\
    & \# passages & \# vocab & \# passages & \# vocab & \# passages & \# vocab \\
    \midrule
    Labeled data & 3,427 & 17,047 & 48,374 & 46,463 & 12,803 & 47,532 \\
    Unlabeled data & - & - & 96,777 & 81,932 & 8,449 & 34,375 \\
    Labeled + Unlabeled data & - & - & 145,151 & 109,415 & 21,252 & 57,863 \\
    \bottomrule
  \end{tabular}
  \begin{minipage}{\columnwidth}
    \centering
\footnotesize\emph{Note:} There are no unlabeled data in the Orchid dataset due to the lack of the same word segmentation and POS tag set.
\end{minipage}
\end{table}

\begin{table}
  \centering
  \caption{Average number of words per passage that exist in the unlabeled data but not in the labeled data.}
  \label{tab:new_vocab}
  \begin{tabular}{p{2cm}|c}
  \toprule
  Dataset & \# words \\
  \midrule
  UGWC & 0.650 \\
  IWSLT & 1.225 \\
  \bottomrule
  \end{tabular}
\end{table}
\subsubsection{Thai sentence segmentation} \label{sec:thaidata}
\paragraph{Orchid \cite{sornlertlamvanich1997orchid}}
This dataset is a Thai part-of-speech-tagged dataset containing 10,864 sentences. In the corpus, text was separated into paragraphs, sentences, and words hierarchically by linguists. Each word was also manually assigned a POS by linguists. These data include no unlabeled data with the same word segmentation and POS tag set. Hence, we do not execute CVT on this dataset.

Our data preprocessing on the ORCHID corpus was similar to that in \cite{zhou-etal-2016-word}: all the comments are removed, and the data are partitioned into 10 parts containing equal numbers of sentences to support 10-fold cross-validation. Each training set is split into one part used for validation and the rest is used for model training. Subsequently, all the words in each dataset are concatenated and then separated into sequences with 200 words per instance. Each sequence always begins with the first word of a sentence. If a sequence ends with an unfinished sentence, the next sequence starts with that complete sentence.

\paragraph{UGWC (User-Generated Web Content) \cite{lertpiya2019preliminary}}
This Thai dataset includes many types of labeled data useful in sentence segmentation tasks. The raw text was generated by users having conversations in the financial domain and were acquired mainly by crawling social sites. The labeled data for sentence segmentation were manually annotated by linguists using the definitions in \cite{lertpiya2019preliminary}.

At the time of this study, the dataset was extended from that in \cite{lertpiya2019preliminary}; the data were collected from January 2017 to December 2017. The labeled dataset includes 48,374 passages. To support semi-supervised learning, the first 3 months of data (96,777 passages) are unlabeled.

Because the data stem from social media, some text exists that cannot be considered as part of any sentence, such as product links, symbols unrelated to sentences, and space between sentences. These portions were not originally annotated as sentences by the linguists. However, in this work, we treat these portions as individual sentences and tag the first word of each fraction as the sentence boundary.

For evaluation purposes, the collection of passages in this dataset is based on 5-fold cross-validation, similar to the previous work \cite{lertpiya2019preliminary}. The passages are treated as input sequences for the model. For each passage, word segmentation and POS tagging are processed by the custom models from this dataset.

\subsubsection{English punctuation restoration}
\paragraph{IWSLT \cite{federico2012overview}}
We adopted this English-language dataset to enable comparisons with models intended for other languages. The dataset is composed of TED talk transcripts. To compare our model with those of previous works, we selected the training dataset for the machine translation track in IWSLT2012 and separated it into training and validation sets containing 2.1 million and 295 thousand words, respectively. The testing dataset is the IWSLT2011 reference set, which contains 13 thousand words. To acquire unlabeled data for semi-supervised learning, we adopted the IWSLT2016 machine translation track training data; duplicate talks that also appear in IWSLT2012 are discarded.

The data preprocessing follows the process in \cite{tilk2016bidirectional}. Each sequence is generated from 200 words, of which beginning is always the first word in a sentence. If a sentence is cut at the end of a sequence, that sentence is copied in full to the beginning of the next sequence.

To use our model, the POS of each word is required. However, the IWSLT dataset contains only the raw text of transcripts and does not include POS tags. Thus, we implement POS tagging using a special library \cite{honnibal2017spacy} to predict the POS of each word.

\subsection{Implementation Detail}

Before mapping each token included in the unigram, bigram, and trigram to the embedding vector, we limit the minimum frequency of occurring words that are not marked as an unknown token. There are 2 parameters set for the unigram $C_{word}$ and the remaining $C_{ngram}$, respectively. We found that model accuracy is highly sensitive to these parameters. Therefore, we use a grid search technique to find the best value for both parameters for the model.

We apply two optimizers used in this work: Adagard \cite{duchi2011adaptive} and Adam \cite{kingma2014adam}, whose learning rates are set to 0.02 and 0.001 for the Thai and English datasets, respectively. To generalize the model, we also integrate L2 regularization with an alpha of 0.01 to the loss function for model updating. Moreover, dropout is applied to the local representation vectors, recurrent representation vectors, between all bidirectional LSTMs and enclosed by the self-attention mechanism in the high-level module.

During training, both the supervised and semi-supervised models are trained until the validation metrics stop improving; the metrics are (1) sentence boundary F1 score and (2) overall F1 score for Thai sentence segmentation and English punctuation restoration, respectively.

CVT has three main parameters that impact model accuracy. The first is the drop rate of the masked language model, which determines the number of tokens that are dropped and used for learning auxiliary prediction modules as described in Section~\ref{cross-view}. The second is the number of unlabeled mini-batches $B$ used for training between supervised mini-batches. Third, rather than using the same dropout rate for the local representation vectors, a new dropout rate is assigned.

The details of hyperparameters such as the hidden size of each layer and dropout rate are given in Section~\ref{appendix:hyperparameters}.

\subsection{Evaluation}
During the evaluation, each task is assessed using different metrics based on previous works. For Thai sentence segmentation, three metrics are used in the evaluation: sentence boundary F1 score, non-sentence boundary F1 score, and space correct \cite{zhou-etal-2016-word}. In this work, we mainly focus on the performance of sentence boundary prediction and not non-sentence boundary prediction or space prediction. Therefore, we make comparisons with other models regarding only their sentence boundary F1 scores. The equation for the sentence boundary F1 score metric is shown in \Cref{f1sb}. In calculating the F1 score, the positive class is defined as the sentence boundary, and the negative class is defined as the non-sentence boundary.

\begin{displaymath}
  precision_{sb} = \frac{\# \textrm{Collectly predicted sentence boundaries}}{\# \textrm{All predicted sentence boundaries}}
\end{displaymath}
\begin{displaymath}
  recall_{sb} = \frac{\# \textrm{Collectly predicted sentence boundaries}}{\# \textrm{All expected sentence boundaries}}
\end{displaymath}
\begin{equation}
  F_{1,sb} = \frac{2 \times precision_{sb} \times recall_{sb}}{precision_{sb} + recall_{sb}}
  \label{f1sb}
\end{equation}

For English punctuation, the evaluation is measured on each type of punctuation and overall F1 score. For the punctuation restoration task, we care only about the performance of the samples belonging to the classes that are tagged to words followed by punctuation; therefore class $O$, which represents words not immediately followed by punctuation, is ignored in the evaluation. Consequently, the overall F1 score does not include $O$ as the positive class in \Cref{f1overall}.

\begin{displaymath}
  precision_{overall} = \frac{\# \textrm{Collectly predicted punctuation marks}}{\# \textrm{All predicted punctuation marks}}
\end{displaymath}
\begin{displaymath}
  recall_{overall} = \frac{\# \textrm{Collectly predicted punctuation marks}}{\# \textrm{All expected punctuation marks}}
\end{displaymath}
\begin{equation}
  F_{1,overall} = \frac{2 \times precision_{overall} \times recall_{overall}}{precision_{overall} + recall^{overall}}
  \label{f1overall}
\end{equation}

To compare the performance of each punctuation restoration model in a manner similar to sentence segmentation, the 2-class F1 score is calculated to measure model accuracy, as shown in \Cref{f12class}. The calculation of this metric is the same as that used in \cite{che2016punctuation}. The metric considers only where the punctuation position is and ignores the type of restored punctuation. Therefore, this measure is similar to the metric sentence boundary F1, which only considers the position of the missing punctuation.

\begin{displaymath}
  precision_{2-class} = \frac{\# \textrm{Collectly predicted punctuation positions}}{\# \textrm{All predicted punctuation positions}}
\end{displaymath}
\begin{displaymath}
  recall_{2-class} = \frac{\# \textrm{Collectly predicted punctuation positions}}{\# \textrm{All expected punctuation positions}}
\end{displaymath}
\begin{equation}
  F_{1,2-class} = \frac{2 \times precision_{2-class} \times recall_{2-class}}{precision_{2-class} + recall_{2-class}}
  \label{f12class}
\end{equation}

\section{Results and discussions} \label{sec:result}
We report and discuss the results of our two tasks in four subsections. The first and second subsections include the effect of local representation and distant representation, respectively. The impact of CVT is explained in the third subsection. The last subsection presents a comparison of our model and all the baselines. Moreover, we also conduct paired t-tests to investigate the significance of the improvement from each contribution, as shown in Section~\ref{sec:ttest}.

\subsection{Effect of local representation}
To find the effect of local representation, we compare a standard Bi-LSTM-CRF model using our full implementation to the model that includes n-gram embedding to extract local representation. In \Cref{tab:thairesult,tab:engresult}, the standard Bi-LSTM-CRF model is represented as Bi-LSTM-CRF (row (e)), while the models with local features are represented as $+ local$ (row (f)).

\paragraph{Thai sentence segmentation}
The results in Table~\ref{tab:thairesult} show that using n-gram to obtain the local representation improves the F1 score of the model from 90.9\% (row (e)) to 92.4\% (row (f)) on the Orchid dataset and from 87.6\% (row (e)) to 88.7\% (row (f)) on the UGWC dataset. These results occur because many word groups exist that can be used to signal the beginning and end of a sentence in Thai. Word groups always found near sentence boundaries can be categorized into 2 groups. The first group consists of final particles, e.g., ``\foreignlanguage{thai}{นะ|คะ} '' (na | kha), ``\foreignlanguage{thai}{นะ|ครับ} '' (na | khrạb), ``\foreignlanguage{thai}{เลย|ครับ} '' (ley | khrạb), ``\foreignlanguage{thai}{แล้ว|ครับ} '' (la{\^e}w | khrạb), and others. These word groups are usually used at the ends of sentences to indicate the formality level. For instance, the model with local representation can detect the sentence boundary at ``\foreignlanguage{thai}{ครับ} '' (khrạb) that is followed by ``\foreignlanguage{thai}{แล้ว} '' (la{\^e}w), as shown in Fig.~\ref{fig:cpngram}, while the model without local representation cannot detect the word as a sentence boundary. The second group consists of conjunctions that are always used at the beginnings of sentences, e.g., ``\foreignlanguage{thai}{จาก|นั้น} (after that) '', ``\foreignlanguage{thai}{ไม่|งั้น} (otherwise) '' and others. The model that uses n-gram to capture word group information is better able to detect word groups near sentence boundaries. Thus, this model can identify these sentence boundaries easily in the Thai language.

\begin{table}
  \caption{The result of Thai sentence segmentation for each model. For the Orchid dataset, we report the average of each metric on 10-fold cross-validation. Meanwhile, average metrics from 5-fold cross-validation are shown for the UGWC dataset.
  }
  \label{tab:thairesult}
  \begin{minipage}{\columnwidth}
  \begin{center}
  \begin{tabular}{p{4cm}|ccc|ccc}
  \toprule
	\multirow[t]{3}{*}{Model} & 
	\multicolumn{3}{c|}{Orchid}  &
	\multicolumn{3}{c}{UGWC} 
	\\
	& precision & recall & F1
	& precision & recall & F1
	\\
  \midrule
	(a) POS-trigram \cite{mittrapiyanuruk2000automatic} & 74.4 & 79.8 & 77.0 & - & - & - \\ 
	(b) Winnow \cite{charoenpornsawat2001automatic} & 92.7 & 77.3 & 84.3 & - & - & - \\ 
	(c) ME \cite{slayden2010thai} & 86.2 & 83.5 & 84.8 & - & - & - \\ 
	(d) CRF (Thai baseline) \cite{zhou-etal-2016-word} & 94.7 & 89.3 & 91.9 & 87.4 & 82.7 & 85.0 \\ 
	 \hline
  (e) BI-LSTM-CRF \cite{huang2015bidirectional} & 92.1 & 89.7 & 90.9 & 87.8 & 87.4 & 87.6 \\ 
  (f) + local & 93.1 & 91.7 & 92.4 & 88.4 & 89.0 & 88.7 \\
  (g) + local + distant & \textbf{93.5} & \textbf{91.5} & \textbf{92.5} & 88.8 & 88.8 & 88.8 \\
  (h) + local + distant + CVT & - & - & - & \textbf{88.9} & \textbf{89.0} & \textbf{88.9} \\
  \bottomrule
\end{tabular}
\end{center}
\footnotesize\emph{Note:} The CVT model is not tested on the Orchid dataset because of the lack of unlabeled data.
\end{minipage}
\end{table}

\begin{figure}[!b]
    \centering
    \begin{tikzpicture}
        \node[text centered] at (2, 2) {Bi-LSTM-CRF};
        \node[text centered] at (2, 1) {+ local};
        \foreach \text \x in {\foreignlanguage{thai}{มี}/4, \foreignlanguage{thai}{นาน}/5, \foreignlanguage{thai}{แล้ว}/6, \foreignlanguage{thai}{ครับ}/7, \foreignlanguage{thai}{เริ่มต้น}/8.25, \foreignlanguage{thai}{ที่}/9.25, \foreignlanguage{thai}{1}/10.25, \foreignlanguage{thai}{ล้าน}/11.25}
            {\node[text centered, anchor=north, text height=0.25] at (\x, 3.5) {\text};}
        \foreach \text \x in {(mī)/4, (nān)/5, (la{\^e}w)/6, (khrạb)/7, (re{\`i}m {\^t}n)/8.25, (thī)/9.25, ({\=h}n{\'{\d{u}}}ng)/10.25, ({\^l}ān)/11.25}
            {\node[text centered, anchor=north, text height=0.25] at (\x, 3) {\text};}
        \draw (0.5,2.5) -- (12.5,2.5);
        \foreach \text \x in {sb/11.5}
            {\node[text centered, anchor=north, text height=0.25] at (\x, 2) {\text};}
        \draw (0.5,1.5) -- (12.5,1.5);
        \foreach \text \x in {sb/7, sb/11.5}
            {\node[text centered, anchor=north, text height=0.25] at (\x, 1) {\text};}
    \end{tikzpicture}
\caption{
    An example of sentence boundary prediction by a normal Bi-LSTM-CRF and by the model with local representation ($+local$). Here, $sb$ indicates that the word is predicted as the sentence boundary.    
}
\label{fig:cpngram}
\end{figure}

\paragraph{English punctuation restoration}
In contrast, for the English dataset, local representation using n-gram drops the overall F1 score of punctuation restoration from 64.4\% (row (e)) to 63.6\% (row (f)), as shown in Table~\ref{tab:engresult}. However, the 2-class F1 score increases slightly from 81.4\% (row (e)) to 81.8\% (row (f)) when compared to the Bi-LSTM-CRF model, which does not integrate n-gram embedding. Common phrases such as ''In spite of'', ''Even though'' and ''Due to the fact'' might provide strong cues for punctuation; however, such phrases can be found at both the beginnings and in the middle of sentences. Because such phrases can be used in both positions, they may follow commas when they are in the middle of the sentence or periods when they are at the beginning of a sentence. However, they still follow either a period or a comma; consequently, such phrases can still help identify whether the punctuation should be restored, which increases the 2-class F1 score, which considers only the positions of missing punctuation. Moreover, English does not use the concept of a final particle usually found at the end of the sentence—similar to the Thai word group mentioned earlier—including ``\foreignlanguage{thai}{นะ|คะ} ''(na | kha), ``\foreignlanguage{thai}{นะ|ครับ} ''(na | khrạb), ``\foreignlanguage{thai}{เลย|ครับ} '' (ley|khrạb), ``\foreignlanguage{thai}{แล้ว|ครับ} '' (la{\^e}w | khrạb) and others. Therefore, the word groups captured by n-gram can only help to identify where punctuation should be restored but they do not help the model determine the type of punctuation that should be restored.

\begin{table}
  \caption{The result of English punctuation restoration by each model. Each model is evaluated by the F1 score of each class of punctuation, the overall F1 score and the 2-class F1 score.}
  \label{tab:engresult}
  \begin{minipage}{\columnwidth}
  \begin{center}
  \begin{tabular}{p{4cm}|ccc|cc}
    \toprule
	Model & comma	& period	& question mark & overall & 2-class
	\\
  \midrule
    (a) T-LSTM \cite{tilk2015lstm} & 45.1 & 56.6 & 49.4 & 50.8 & - \\
    (b) T-BRNN \cite{tilk2016bidirectional} & 53.1 & 71.9 & 62.8 & 63.1 & - \\
    (c) T-BRNN-pre \cite{tilk2016bidirectional} & 54.8 & 72.9 & 66.7 & 64.4 & - \\
    (d) CRF (Thai baseline) \cite{zhou-etal-2016-word} & 44.9 & 60.8 & 26.7 & 52.7 & - \\
	 \hline
	  (e) BI-LSTM-CRF \cite{huang2015bidirectional} & 55.3 & 73.1 & 63.5 & 64.4 & 81.4 \\
    (f) + local & 55.0 & 72.3 & 63.3 & 63.6 & 81.8 \\
    (g) + local + distant & \textbf{56.7} & 72.9 & 59.2 & 64.5 & 81.7 \\
    (h) + local + distant + CVT & 56.1 & \textbf{73.8} & \textbf{66.6} & \textbf{65.2} & \textbf{82.7} \\
  \bottomrule
\end{tabular}
\end{center}
\footnotesize\emph{Note:} F1 score for the ''O'' class, which indicates that there is no punctuation following the word, is not calculated because we care only the performance of each punctuation class.
\end{minipage}
\end{table}

\subsection{Effect of distant representation}
The effect of this contribution can be found by comparing the model that integrates the distant representation and the model that does not. The model with distant features integrated is represented as $+ local + distant$ (row (g)) in both tables. In this case, the distant representation is composed of the self-attention modules in both the low- and high-level modules, as shown in Fig.~\ref{fig:architecture}.

From the combination of local and distant representation, the results in \Cref{tab:thairesult,tab:engresult} show that the distant feature improves the accuracy of the model on all datasets compared to the model with no distant representation. The F1 scores of the sentence segmentation models improved slightly, from 92.4\% and 88.7\% (row (f)) to 92.5\% and 88.8\% (row (g)) on the Orchid and UGWC datasets, respectively. For the IWSLT dataset, the distant feature can recover the overall F1 score of punctuation restoration, which is degraded by the n-gram embedding; it improves from 63.6\% (row (f)) to 64.5\% (row (g)). The reason is that the self-attention modules focus selectively on certain parts of the passage. Thus, the model focuses on the initial words of the dependent clauses, which helps in classifying which type of punctuation should be restored. An example is shown in Fig.~\ref{fig:cpdistant}: the model with distant representation classifies the punctuation after ''her'' as a ''COMMA'' because ''Before'' is the word that indicates the dependent clause. Meanwhile, the model without distant representation predicts the punctuation as a ''PERIOD'' because there is no self-attention module; therefore, it does not focus on the word ''Before''. Overall, the model that includes both local and distant representation can generally be used for both sentence segmentation and punctuation restoration, and it outperforms both baseline models.

\begin{figure}
    \centering
    \begin{tikzpicture}
        \node[text centered] at (2, 2) {+ local};
        \node[text centered] at (2, 1) {+ local + distant};
        \foreach \text \x in {Before/4, I/5, met/6, her/7, I/8, went/9, through/10, suboptimal/11.5, search/13, results/14}
            {\node[text centered, anchor=north, text height=0.25] at (\x, 3) {\text};}
        \draw (0.5,2.5) -- (14.5,2.5);
        \foreach \text \x in {PERIOD/7, PERIOD/14}
            {\node[text centered, anchor=north, text height=0.25] at (\x, 2) {\text};}
        \draw (0.5,1.5) -- (14.5,1.5);
        \foreach \text \x in {COMMA/7, PERIOD/14}
            {\node[text centered, anchor=north, text height=0.25] at (\x, 1) {\text};}
    \end{tikzpicture}
\caption{
    An example of punctuation prediction of the model with distant representation ($+\ local\ +\ distant$) and without distant representation ($+\ local$). The "COMMA" indicates that the word is followed by a comma (,) and "PERIOD" indicates that a period (.) is restored at that position. }
\label{fig:cpdistant}
\end{figure}

\subsection{Effect of Cross-View Training (CVT)} \label{sec:cross-view}

To identify the improvement from CVT, we compared the models that use different training processes: standard supervised training ($+\ local\ +\ distant$) and CVT ($+\ local\ +\ distant\ +\ CVT$). The model trained with CVT improves the accuracy in terms of the F1 score on both Thai and English datasets, as shown in \Cref{tab:thairesult,tab:engresult} (row (g) vs row (h)).

\paragraph{Thai sentence segmentation}
This experiment was conducted only on the UGWC dataset because no unlabeled data are available in the Orchid dataset, as mentioned in Section~\ref{sec:thaidata}. The model improves the F1 score slightly, from 88.8\% (row (g)) to 88.9\% (row (h)) on the UGWC dataset. This result occurs because both the labeled and unlabeled data in the UGWC dataset are drawn from the same finance domain. The average number of new words found in a new unlabeled data passage is only 0.650, as shown in Table~\ref{tab:new_vocab}. Therefore, there is little additional information to be learned from unlabeled data.

\paragraph{English punctuation restoration}
CVT also improved the model on the IWSLT dataset, from an overall F1 score of 64.5\% (row (g)) to 65.3\% (row (h)) and from a 2-class F1 score of 81.7\% to 82.7\%. Because both the labeled and unlabeled data were collected from TED talks, the number of vocabulary words grows substantially more than in the UGWC dataset because the talks cover various topics. In this dataset, average 1.225 new words found in each new unlabeled data passage, as shown in Table~\ref{tab:new_vocab}; consequently the model representation learns new information from these new words effectively.


\subsection {Comparison with baseline models}
\paragraph{Thai sentence segmentation}
For the Thai sentence segmentation task, our model is superior to all the baselines on both Thai sentence segmentation datasets, as shown in Table~\ref{tab:thairesult}. On the Orchid dataset, the supervised model that includes both local and distant representation was adopted for comparison to the baseline model. Our model improves the F1 score achieved by CRF-ngram, which is the state-of-the-art model for Thai sentence segmentation in Orchid, from 91.9\% (row (d)) to 92.5\% (row (g)). Meanwhile, in the UGWC dataset, our CVT model (row (h)) achieves an F1 score of 88.9\%, which is higher than the F1 score of both the baselines (CRF-ngram and Bi-LSTM-CRF (rows d and e, respectively)). Thus, our model is now the state-of-the-art model for Thai sentence segmentation on both the Orchid and UGWC datasets.

\paragraph{English punctuation restoration}
Our model outperforms all the sequence tagging models. T-BRNN-pre (row (c)) is the current state-of-the-art model, as shown in Table~\ref{tab:engresult}. The CVT model improves the overall F1 score from the 64.4\% of T-BRNN-pre to 65.3\% (row (h)), despite the fact that T-BRNN-pre integrates a pretrained word vector. Moreover, our model also achieves a 2-class F1 score 1.3\% higher than that of Bi-LSTM-CRF (row (e)).

\section{Conclusions} \label{sec:conclusion}
In this paper, we propose a novel deep learning model for Thai sentence segmentation. This study makes three main contributions. The first contribution is to integrate a local representation based on n-gram embedding into our deep model. This approach helps to capture word groups near sentence boundaries, allowing the model to identify boundaries more accurately. Second, we integrate a distant representation obtained from self-attention modules to capture sentence contextual information. This approach allows the model to focus on the initial words of dependent clauses (i.e., ''Before'', ''If'', and ''Although''). The last contribution is an adaptation of CVT, which allows the model to utilize unlabeled data to produce effective local and distant representations.

The experiment was conducted on two Thai datasets, Orchid and UGWC, and one English punctuation restoration dataset, IWSLT. English punctuation restoration is similar to our Thai sentence segmentation. On the Thai sentence segmentation task, our model achieves F1 scores of 92.5\% and 88.9\% on the Orchid and UGWC datasets, constituting a relative error reduction of 7.4\% and 10.5\%, respectively. On the English punctuation task, the 2-class F1 score reached 82.7\% when considering only two punctuation classes (making the task similar to sentence segmentation in Thai). Moreover, our model outperforms the model integrated with pretrained word vectors in terms of the overall F1 score on the IWSLT dataset. Based on our contributions, the local representation scheme has the highest impact on the Thai corpus, while the distant representation and CVT result in strong improvements on the English dataset.

Moreover, our model can also be applied to elementary discourse unit (EDU) segmentation, which is used as the minimal syntactic unit for downstream tasks such as text summarization and machine translation. However, no experiments have been conducted to determine how different sentences and EDUs affect downstream tasks. Therefore, the evaluation of downstream tasks from different sources needs to be studied.

\section*{Acknowledgment}
This paper was supported by KLabs at Kasikorn Business Technology (KBTG), who provided facilities and data. The procedures that were conducted based on social data are visible to the public, and ethical issues that can arise from the use of such data were addressed. We would like to thank the linguists Sasiwimon Kalunsima, Nutcha Tirasaroj, Tantong Champaiboon and Supawat Taerungruang for annotating the UGWC dataset used in this study.

\bibliographystyle{amsplain}
\bibliography{sample-base}

\providecommand{\bysame}{\leavevmode\hbox to3em{\hrulefill}\thinspace}
\providecommand{\MR}{\relax\ifhmode\unskip\space\fi MR }
\providecommand{\MRhref}[2]{%
  \href{http://www.ams.org/mathscinet-getitem?mr=#1}{#2}
}
\providecommand{\href}[2]{#2}
\begin{thebibliography}{10}

\bibitem{afsharizadeh2018query}
Mahsa Afsharizadeh, Hossein Ebrahimpour-Komleh, and Ayoub Bagheri,
  \emph{Query-oriented text summarization using sentence extraction technique},
  2018 4th International Conference on Web Research (ICWR), IEEE, 2018,
  pp.~128--132.

\bibitem{aroonmanakun2007thoughts}
Wirote Aroonmanakun et~al., \emph{Thoughts on word and sentence segmentation in
  thai}, Proceedings of the Seventh Symposium on Natural language Processing,
  Pattaya, Thailand, December 13--15, 2007, pp.~85--90.

\bibitem{charoenpornsawat2001automatic}
Paisarn Charoenpornsawat and Virach Sornlertlamvanich, \emph{Automatic sentence
  break disambiguation for thai}, International Conference on Computer
  Processing of Oriental Languages (ICCPOL), 2001, pp.~231--235.

\bibitem{che2016punctuation}
Xiaoyin Che, Cheng Wang, Haojin Yang, and Christoph Meinel, \emph{Punctuation
  prediction for unsegmented transcript based on word vector}, Proceedings of
  the Tenth International Conference on Language Resources and Evaluation (LREC
  2016), 2016, pp.~654--658.

\bibitem{cho2015punctuation}
Eunah Cho, Jan Niehues, Kevin Kilgour, and Alex Waibel, \emph{Punctuation
  insertion for real-time spoken language translation}, Proceedings of the
  Eleventh International Workshop on Spoken Language Translation, 2015.

\bibitem{clark2018semi}
Kevin Clark, Minh-Thang Luong, Christopher~D Manning, and Quoc Le,
  \emph{Semi-supervised sequence modeling with cross-view training},
  Proceedings of the 2018 Conference on Empirical Methods in Natural Language
  Processing, 2018, pp.~1914--1925.

\bibitem{devlin2018bert}
Jacob Devlin, Ming-Wei Chang, Kenton Lee, and Kristina Toutanova, \emph{Bert:
  Pre-training of deep bidirectional transformers for language understanding},
  arXiv preprint arXiv:1810.04805 (2018).

\bibitem{duchi2011adaptive}
John Duchi, Elad Hazan, and Yoram Singer, \emph{Adaptive subgradient methods
  for online learning and stochastic optimization}, Journal of Machine Learning
  Research \textbf{12} (2011), no.~Jul, 2121--2159.

\bibitem{federico2012overview}
Marcello Federico, Mauro Cettolo, Luisa Bentivogli, Paul Michael, and
  St{\"u}ker Sebastian, \emph{Overview of the iwslt 2012 evaluation campaign},
  IWSLT-International Workshop on Spoken Language Translation, 2012,
  pp.~12--33.

\bibitem{gillick2009sentence}
Dan Gillick, \emph{Sentence boundary detection and the problem with the us},
  Proceedings of Human Language Technologies: The 2009 Annual Conference of the
  North American Chapter of the Association for Computational Linguistics,
  Companion Volume: Short Papers, 2009, pp.~241--244.

\bibitem{gotoh2000sentence}
Yoshihiko Gotoh and Steve Renals, \emph{Sentence boundary detection in
  broadcast speech transcripts}, ASR2000-Automatic Speech Recognition:
  Challenges for the new Millenium ISCA Tutorial and Research Workshop (ITRW),
  2000.

\bibitem{gravano2009restoring}
Agustin Gravano, Martin Jansche, and Michiel Bacchiani, \emph{Restoring
  punctuation and capitalization in transcribed speech}, 2009 IEEE
  International Conference on Acoustics, Speech and Signal Processing, IEEE,
  2009, pp.~4741--4744.

\bibitem{hochreiter1997long}
Sepp Hochreiter and J{\"u}rgen Schmidhuber, \emph{Long short-term memory},
  Neural computation \textbf{9} (1997), no.~8, 1735--1780.

\bibitem{honnibal2017spacy}
Matthew Honnibal and Ines Montani, \emph{spacy 2: Natural language
  understanding with bloom embeddings, convolutional neural networks and
  incremental parsing}, To appear (2017).

\bibitem{huang2015bidirectional}
Zhiheng Huang, Wei Xu, and Kai Yu, \emph{Bidirectional lstm-crf models for
  sequence tagging}, arXiv preprint arXiv:1508.01991 (2015).

\bibitem{jacovi2018understanding}
Alon Jacovi, Oren~Sar Shalom, and Yoav Goldberg, \emph{Understanding
  convolutional neural networks for text classification}, arXiv preprint
  arXiv:1809.08037 (2018).

\bibitem{kingma2014adam}
Diederik~P Kingma and Jimmy Ba, \emph{Adam: A method for stochastic
  optimization}, arXiv preprint arXiv:1412.6980 (2014).

\bibitem{kolavr2012development}
J{\'a}chym Kol{\'a}{\v{r}} and Lori Lamel, \emph{Development and evaluation of
  automatic punctuation for french and english speech-to-text}, Thirteenth
  Annual Conference of the International Speech Communication Association,
  2012.

\bibitem{kolavr2006using}
J{\'a}chym Kol{\'a}{\v{r}}, Elizabeth Shriberg, and Yang Liu, \emph{Using
  prosody for automatic sentence segmentation of multi-party meetings},
  International Conference on Text, Speech and Dialogue, Springer, 2006,
  pp.~629--636.

\bibitem{lafferty2001conditional}
John Lafferty, Andrew McCallum, and Fernando~CN Pereira, \emph{Conditional
  random fields: Probabilistic models for segmenting and labeling sequence
  data},  (2001).

\bibitem{lertpiya2019preliminary}
Anuruth Lertpiya, Teerapat Chaiwachirasak, Nattasit Maharattanamalai, Theerapat
  Lapjaturapit, Tawunrat Chalothorn, Nutcha Tirasaroj, and Ekapol
  Chuangsuwanich, \emph{A preliminary study on fundamental thai nlp tasks for
  user-generated web content}, 2018 International Joint Symposium on Artificial
  Intelligence and Natural Language Processing (iSAI-NLP), IEEE, 2019,
  pp.~1--8.

\bibitem{lu2010better}
Wei Lu and Hwee~Tou Ng, \emph{Better punctuation prediction with dynamic
  conditional random fields}, Proceedings of the 2010 conference on empirical
  methods in natural language processing, 2010, pp.~177--186.

\bibitem{mihalcea2004graph}
Rada Mihalcea, \emph{Graph-based ranking algorithms for sentence extraction,
  applied to text summarization}, Proceedings of the ACL Interactive Poster and
  Demonstration Sessions, 2004.

\bibitem{mittrapiyanuruk2000automatic}
Pradit Mittrapiyanuruk and Virach Sornlertlamvanich, \emph{The automatic thai
  sentence extraction}, Proceedings of the fourth symposium on Natural Language
  Processing, 2000, pp.~23--28.

\bibitem{miyato2016adversarial}
Takeru Miyato, Andrew~M Dai, and Ian Goodfellow, \emph{Adversarial training
  methods for semi-supervised text classification}, arXiv preprint
  arXiv:1605.07725 (2016).

\bibitem{miyato2018virtual}
Takeru Miyato, Shin-ichi Maeda, Shin Ishii, and Masanori Koyama, \emph{Virtual
  adversarial training: a regularization method for supervised and
  semi-supervised learning}, IEEE transactions on pattern analysis and machine
  intelligence (2018).

\bibitem{peitz2011modeling}
Stephan Peitz, Markus Freitag, Arne Mauser, and Hermann Ney, \emph{Modeling
  punctuation prediction as machine translation}, International Workshop on
  Spoken Language Translation (IWSLT) 2011, 2011.

\bibitem{raiman2017globally}
Jonathan Raiman and John Miller, \emph{Globally normalized reader}, arXiv
  preprint arXiv:1709.02828 (2017).

\bibitem{rasmus2015semi}
Antti Rasmus, Mathias Berglund, Mikko Honkala, Harri Valpola, and Tapani Raiko,
  \emph{Semi-supervised learning with ladder networks}, Advances in neural
  information processing systems, 2015, pp.~3546--3554.

\bibitem{ruder2018strong}
Sebastian Ruder and Barbara Plank, \emph{Strong baselines for neural
  semi-supervised learning under domain shift}, arXiv preprint arXiv:1804.09530
  (2018).

\bibitem{slayden2010thai}
Glenn Slayden, Mei-Yuh Hwang, and Lee Schwartz, \emph{Thai sentence-breaking
  for large-scale smt}, Proceedings of the 1st Workshop on South and Southeast
  Asian Natural Language Processing, 2010, pp.~8--16.

\bibitem{sornlertlamvanich1997orchid}
Virach Sornlertlamvanich, Thatsanee Charoenporn, and Hitoshi Isahara,
  \emph{Orchid: Thai part-of-speech tagged corpus}, National Electronics and
  Computer Technology Center Technical Report (1997), 5--19.

\bibitem{taylor1953cloze}
Wilson~L Taylor, \emph{“cloze procedure”: A new tool for measuring
  readability}, Journalism Bulletin \textbf{30} (1953), no.~4, 415--433.

\bibitem{tilk2015lstm}
Ottokar Tilk and Tanel Alum{\"a}e, \emph{Lstm for punctuation restoration in
  speech transcripts}, Sixteenth annual conference of the international speech
  communication association, 2015.

\bibitem{tilk2016bidirectional}
\bysame, \emph{Bidirectional recurrent neural network with attention mechanism
  for punctuation restoration.}, 2016.

\bibitem{ueffing2013improved}
Nicola Ueffing, Maximilian Bisani, and Paul Vozila, \emph{Improved models for
  automatic punctuation prediction for spoken and written text.}, Interspeech,
  2013, pp.~3097--3101.

\bibitem{vaswani2017attention}
Ashish Vaswani, Noam Shazeer, Niki Parmar, Jakob Uszkoreit, Llion Jones,
  Aidan~N Gomez, {\L}ukasz Kaiser, and Illia Polosukhin, \emph{Attention is all
  you need}, Advances in neural information processing systems, 2017,
  pp.~5998--6008.

\bibitem{wang2018self}
Feng Wang, Wei Chen, Zhen Yang, and Bo~Xu, \emph{Self-attention based network
  for punctuation restoration}, 2018 24th International Conference on Pattern
  Recognition (ICPR), 2018.

\bibitem{wathabunditkul2003spacing}
Suphawut Wathabunditkul, \emph{Spacing in the thai language}, 2003.

\bibitem{zhou-etal-2016-word}
Nina Zhou, AiTi Aw, Nattadaporn Lertcheva, and Xuancong Wang, \emph{A word
  labeling approach to {T}hai sentence boundary detection and {POS} tagging},
  Proceedings of {COLING} 2016, the 26th International Conference on
  Computational Linguistics: Technical Papers (Osaka, Japan), The COLING 2016
  Organizing Committee, December 2016, pp.~319--327.

\bibitem{zhu2005semi}
Xiaojin~Jerry Zhu, \emph{Semi-supervised learning literature survey}, Tech.
  report, University of Wisconsin-Madison Department of Computer Sciences,
  2005.

\end{thebibliography}

\appendix
\section*{APPENDIX}

\section{Hyperparameters} \label{appendix:hyperparameters}
The hyperparameter values were determined through a grid search to find their optimal values on the different datasets. All the hyperparameters for each dataset are shown in Table~\ref{tab:hyper}. The optimal values from the grid search depend on the task. For Thai sentence segmentation, the hyperparameters are tuned to obtain the highest sentence boundary F1 score, while the overall F1 score is used to tune the parameters for English punctuation restoration.

\begin{table}[!b]
  \caption{Model hyperparameters for each dataset}
  \label{tab:hyper}
  \begin{minipage}{\columnwidth}
  \begin{center}
  \begin{tabular}{p{9cm}|c|c|c}
    \toprule
	Model & Orchid	& UGWC	& IWSLT
	\\
  \midrule
    $C_{word}$ & 2 & 2 & 2 \\
    $C_{ngram}$ & 2 & 2 & 13 \\
    Optimizer & AdaGrad & AdaGrad & Adam \\
    Learning rate & 0.02 & 0.02 & 0.001 \\
    Batch size & 16 & 16 & 16 \\
    Early stopping patience & 5 & 5 & 5 \\
    Unigram embedding size (Text) & 64 & 64 & 300 \\
    Unigram embedding size (POS and Type) & 32 & 32 & 300 \\
    Bigram \& Trigram embedding size (Text) & 16 & 16 & 10 \\
    Bigram \& Trigram embedding size (POS and Type) & 8 & 8 & 10 \\
    LSTM hidden size & 25 & 25 & 256 \\
    Number of LSTM layers in High-level module ($K$) & 2 & 2 & 4 \\
    Self-attention output size & 50 & 50 & 256 \\
    Low-level Self-attention projection size & 64 & 64 & 32 \\
    High-level Self-attention projection size & 25 & 25 & 128 \\
    Local embedding dropout & 0.30 & 0.30 & 0.30 \\
    Dropout between layers & 0.15 & 0.15 & 0.15 \\

    Dropped rate of masked language model & - & 0.30 & 0.30 \\
    Number of unlabeled mini-batch $B$& - & 1 & 2 \\
    Dropout of the unlabeled input & - & 0.50 & 0.30 \\
  \bottomrule
\end{tabular}
\end{center}
\end{minipage}
\end{table}

\section{Comparison of CNN and n-gram models for local representation} \label{appendix:cnn}
Jacovi A. et al. \cite{jacovi2018understanding} proposed that a CNN can be used as an n-gram detector to capture local text features. Therefore, we also performed an experiment to compare a CNN and n-gram embedded as local structures. The results in Table~\ref{tab:cnn} show that the model using the embedded n-gram yields greater improvement than the one using an embedded CNN on the Orchid and UGWC datasets.

\begin{table}[!h]
  \caption{Comparison between CNN and n-gram embedding for local representation extraction}
  \label{tab:cnn}
  \begin{minipage}{\columnwidth}
  \begin{center}
  \begin{tabular}{p{4cm}|c|c}
  \toprule
	Model & ORCHID & UGWC
	\\
  \midrule
    Bi-LSTM-CRF & 90.9\% & 87.6\% \\ 
    Bi-LSTM-CRF + n-gram & 92.4\% & 88.7\% \\
    Bi-LSTM-CRF + CNN & 91.2\% & 87.9\% \\
  \bottomrule
\end{tabular}
\end{center}
\end{minipage}
\end{table}

\section{Statistical Tests for Thai sentence segmentation} \label{sec:ttest}
To prove the significance of the model improvements, we compared the cross-validation results using paired t-tests to obtain the p-values, which are shown in Table~\ref{tab:orchidpvalue} for the Orchid dataset and Table~\ref{tab:ugwcpvalue} for the UGWC dataset.

\begin{table}[!h]
  \caption{The improvement of each contribution on the Orchid dataset results shown as p-values from paired t-tests }
  \label{tab:orchidpvalue}
  \begin{minipage}{\columnwidth}
  \begin{center}
  \begin{tabular}{p{3cm}|c|c|c|c}
  \toprule
	Model & CRF & Bi-LSTM-CRF & + local
	\\
  \midrule
    + local & +0.467\% (0.009) & +1.528\% (<0.001) & -  \\
    + local + distant & +0.454\% (0.002) & +1.596\% (<0.001) & +0.068\% (0.370) \\
  \bottomrule
\end{tabular}
\end{center}
\footnotesize\emph{Note:} The number in the table reflects the percentage of improvement from the columns compared with the rows. The number in parentheses is the p-value computed from a paired t-test.
\end{minipage}
\end{table}

\begin{table}[!h]
  \caption{The improvement of each contribution on the UGWC dataset results shown as p-values from paired t-tests }
  \label{tab:ugwcpvalue}
  \begin{minipage}{\columnwidth}
  \begin{center}
  \begin{tabular}{p{3cm}|c|c|c|c}
  \toprule
	Model & CRF & Bi-LSTM-CRF & + local	& + local + distant
	\\
  \midrule
    local & +3.664\% (<0.001) & +1.085\% (<0.001) & - & - \\
    local + distant & +3.774\% (<0.001) & +1.195\% (<0.001) & +0.110\% (0.182) & - \\
    local + distant + CVT & +3.919\% (<0.001) & +1.340\% (<0.001) & +0.255\% (0.001) & +0.145\% (0.065) \\
  \bottomrule
\end{tabular}
\end{center}
\footnotesize\emph{Note:} The number in the table reflects the percentage of improvement from the columns compared with the rows. The number in parentheses is the p-value computed from a paired t-test.
\end{minipage}
\end{table}

\end{document}